# High-Pass Graph Convolutional Network for Enhanced Anomaly Detection: A Novel Approach

Shelei Li [a,b], Yong Chai Tan[a], Tai Vincent[a]

a Faculty of Engineering, Building Environment and information Technology, SEGi University, Petaling Java, 47810, Malaysia

b School of Information & Intelligence Engineering, University of Sanya, Sanya, 572022, China

**Abstract**

Graph Convolutional Network (GCN) are widely used in Graph Anomaly Detection (GAD) due to their natural compatibility with graph structures, resulting in significant performance improvements. However, most researchers approach GAD as a graph node classification task and often rely on low-pass filters or feature aggregation from neighboring nodes. This paper proposes a novel approach by introducing a High-Pass Graph Convolution Network (HP-GCN) for GAD. The proposed HP-GCN leverages high-frequency components to detect anomalies, as anomalies tend to increase high-frequency signals within the network of normal nodes. Additionally, isolated nodes, which lack interactions with other nodes, present a challenge for Graph Neural Network (GNN). To address this, the model segments the graph into isolated nodes and nodes within connected subgraphs. Isolated nodes learn their features through Multi-Layer Perceptron (MLP), enhancing detection accuracy. The model is evaluated and validated on YelpChi, Amazon, T-Finance, and T-Social datasets. The results showed that the proposed HP-GCN can achieve anomaly detection accuracy of 96.10%, 98.16%, 96.46%,

Li shelei is with is with Faculty of Engineering, Building Environment and information Technology, SEGi University, Pedaling Java, 47810, Malaysia and School of Information & Intelligence Engineering, University of Sanya, Sanya, 572022, China (e-mail: SUKD2202348@segi4u.my).

Yong Chai Tan is with Faculty of Engineering, Building Environment and information Technology, SEGi University, Pedaling Java, 47810, Malaysia (e-mail: tanyongchai@segi.edu.my).

>

and 98.94%, respectively. The findings demonstrate that the HP-GCN outperforms existing GAD methods based on spatial domain GNN as well as those using low-pass and band-pass filters in spectral domain GCN. The findings underscore the effectiveness of this method in improving anomaly detection performance. Source code can be found at: https://github.com/meteor0033/High-pass_GAD.git.

## 1. INTRODUCTION

Anomaly detection seeks to identify objects that significantly deviate from the majority in a dataset [1]. Anomaly detection has attracted considerable attention owing to its broad range of applications, including tasks such as IoT security [2], Spammer Detection [3], Industrial Safety and Monitoring [4][5], financial fraud detection [6][7], and credit card fraud [8]. Many objects inherently exhibit rich interrelationships, which can be intuitively modeled as graphs [9]. Many scholars [10][11][12][13][14] consider using graph neural network (GNN) to conduct graph abnormal nodes detection. owing to their impressive performance in learning representations. However, fundamentally graph anomaly detection typically functions as a semi-supervised classification problem in these models. Using Graph Convolutional Network (GCN) or GNN based on the principle of local smoothness on graphs. However, the number of anomalous nodes is much smaller compared to normal nodes, and anomalous nodes are usually surrounded by a large number of normal nodes. This renders the node characteristics incompatible with the principle of local smoothness on the graph. previous endeavors addressing this issue have focused on improvements in both spatial and spectral domains.

In the spatial domain, Liu et al., [15] introduced GraphConsis model, which examines the context inconsistency, feature inconsistency, and relation inconsistency problems resulting from the aforementioned issues. To mitigate the impact of these inconsistencies on the performance of GNN for anomaly detection, they proposed a consistency score to identify and filter out inconsistent neighbors and learn the attention weights for relations corresponding to these three types of inconsistency. Dou et al., [11] addressed the camouflage behavior of anomalous nodes by designing the CAmouflage-REsistant GNN (CARE-GNN). This model employs three distinct modules: label-aware similarity measurement, selective aggregation of neighbors, and reinforcement learning (RL) to enhance the GNN aggregation process, thereby resisting camouflage attempts. Liu et al., [16] proposed the Pick and Choose Graph Neural Network (PC-GNN) model, building upon the CARE-GNN framework. This model addresses the issue of graph node class imbalance, which is particularly prevalent in domains such as financial fraud and network intrusion. In these areas, the proportion of anomalous nodes is significantly lower than that of normal nodes, resulting in challenges for imbalanced supervised learning on graphs. To tackle this challenge, they devised a label-balanced sampler for node and edge selection to construct mini-batch training subgraphs. Furthermore, they introduced a neighbor sampler to select candidate neighbors for nodes within the subgraph. Finally, they aggregated information from selected neighbors and different relations to obtain the final representation of the target node, thereby alleviating the issue of graph node class imbalance. Gao et al., [17] designed the Graph Decomposition Network (GDN), which aims to address structural distribution shift (SDS) by screening out abnormal features to reduce the impact of heterogeneous features.

In the spectral domain, the BWGNN model recommended aims to 'right-shift' phenomenon of graph spectral induced by anomalous nodes [18]. Inspired by GWNN [19], the model utilizes Beta Wavelet to construct a band-pass filter on the graph, thereby enhancing graph anomaly nodes detection performance to a certain extent. Gao et al., [9] build upon the BWGNN framework to address the sparsity of anomalous nodes, which are often connected to normal nodes. They designed an edge pruning strategy between anomalous and normal nodes to prevent the aggregation of normal node features with anomalous node features. This strategy seeks to maintain the distinctiveness of anomalous node features by avoiding their dilution through similarity with normal node features.

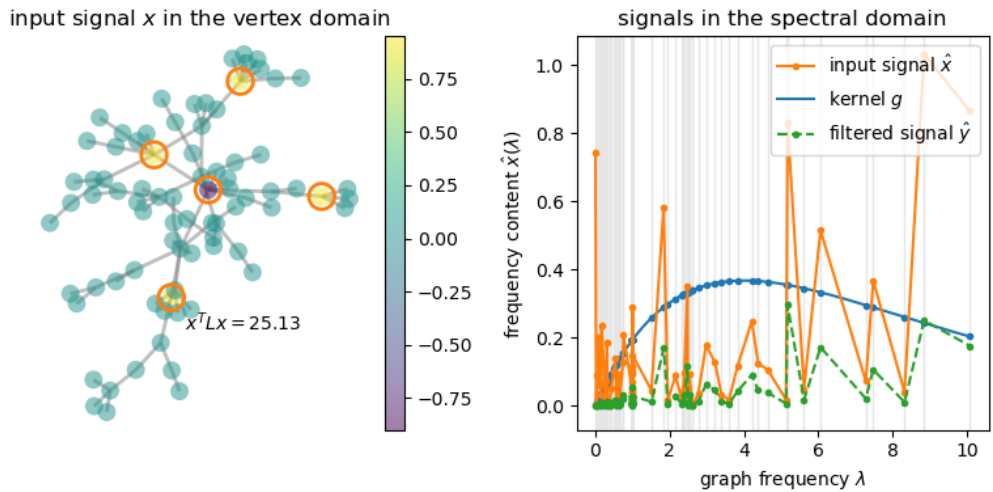

Figure1．Graph signal band-pass filtering

However, in graph anomaly detection problems, anomalies are typically embedded within the network of normal nodes due to the scarcity of anomalous nodes compared to normal ones. This results in an increase in high-frequency components in the spectrum, as illustrated in Figure 1. Utilizing band-pass filters tends to smooth the signal, which is detrimental to the identification of anomalous nodes. This paper utilizes a high-pass filter, which preserves or enhances high-frequency components while attenuating mid-to-low-frequency components, thereby improving the discriminative power of anomalous node features. Additionally, as shown in Figure 2, visualization of the dataset reveals that many peripheral nodes, which include some anomalous nodes, are mostly isolated nodes. These nodes cannot effectively learn feature representations using graph neural network methods. For YelpChi, there are 45954 nodes in YelpChi, of which includes 22123 isolate nodes.

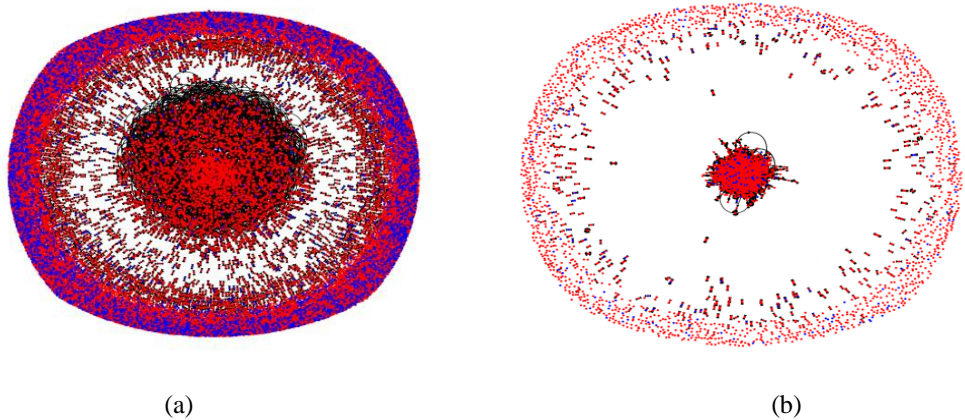

(a)          (b)

(a) Visualization of YelpChi nodes relationship (R-U-R Edges: connect comments authored by the same user) (b) Visualization of Graph Amazon nodes relationship (U-P-U Edges: Connect users who have reviewed at least one common product)

Figure 2. Visualization nodes relationship of Datasets

The key contributions can be outlined as follows:

1) To resolve the increase in high-frequency components caused by anomaly nodes in the graph, a high-pass filter is proposed, enhancing the discriminative features of anomalous nodes. This contributes to the detection of anomalous nodes and provides a fresh perspective for graph node detection problems.

2) To tackle the issue of ineffective learning of node representations caused by numerous isolated nodes in the graph, a method that combines Graph Neural Networks (GNNs) with feature-learning Multilayer Perceptions (MLPs) is proposed. This approach is designed to enhance the detection of anomalous nodes, addressing both isolated nodes and small connected components within the graph.

The rest of this paper is organized as follows: Section 2 provides the mathematical background relevant to graph anomaly detection and high-pass graph filters. Section 3 discusses the high-pass graph convolutional network model and methods for learning features of isolated nodes and small connected components. Section 4 presents and compares the experimental results. Finally, Section 5 offers the paper's conclusions.

**2. PRELIMINARIES**

A. Fundamental Concept

Definition 1 (Attributed Graph), Given an undirected attributed graph $G = \{V, E, X\}$, where $V$ is a finite set of $|V| = n$, that is $V = \{v_i\}_0^{N-1}$, and edge set $E = \{e_{i,j}\}$, where $e_{i,j} = (v_i, v_j)$ represents an edge between nodes $v_i$ and $v_j$. The attribute matrix $X = [x_i]_{N \times K}$, where $x_i$ is a $k$-dimension vector.

Definition 2 (Graph Laplacian）：Let $W$ be the adjacency matrix of graph $G$, and $D \in \mathbb{R}^{n \times n}$ is the diagonal degree matrix with $D_{ii} = \sum_j W_{ij}$. The definition of Graph Laplacian is

$$L = D - W \in \mathbb{R}^{n \times n} \tag{1}$$

and normalized definition is

\>

$$L = I_n - D^{-1/2}WD^{-1/2} \tag{2}$$

Where $I_n$ is the identity matrix. As $L$ is a real symmetric positive semidefinite matrix, it possesses a set of orthogonal eigenvectors, which can serve as a basis for the graph Fourier transform. Its corresponding non-negative eigenvalues are defined as the graph frequencies $\Lambda = \text{diag}([\lambda_0, \cdots, \lambda_{n-1}]) \in \mathbb{R}^{n \times n}$.

Definition 3 (Graph Fourier Transform):

For a signal $x \in \mathbb{R}^n$, its graph Fourier transform can be defined as:

$$\hat{x} = U^T x \tag{3}$$

and its inverse as:

$$x = U\hat{x} \tag{4}$$

B. Graph Filters

Tang et al., [18] addresses the issue of spectral right-shift caused by anomalous nodes by designing a band-pass filter model specifically for detecting anomalous nodes in graphs. However, when the proportion of anomalous nodes is low and submerged among normal nodes, it will cause an increase in high-frequency components. Figure 3 (a) illustrated a case where anomalous nodes are submerged among normal nodes, and different band-pass filters are employed for filtering. Figure 3 (b) shows the result of the signal x in the spectral domain after being processed by the band-pass filter g, resulting in the signal y. Figure 3 (c) presents the graph signal in the spatial domain after filtering.

>

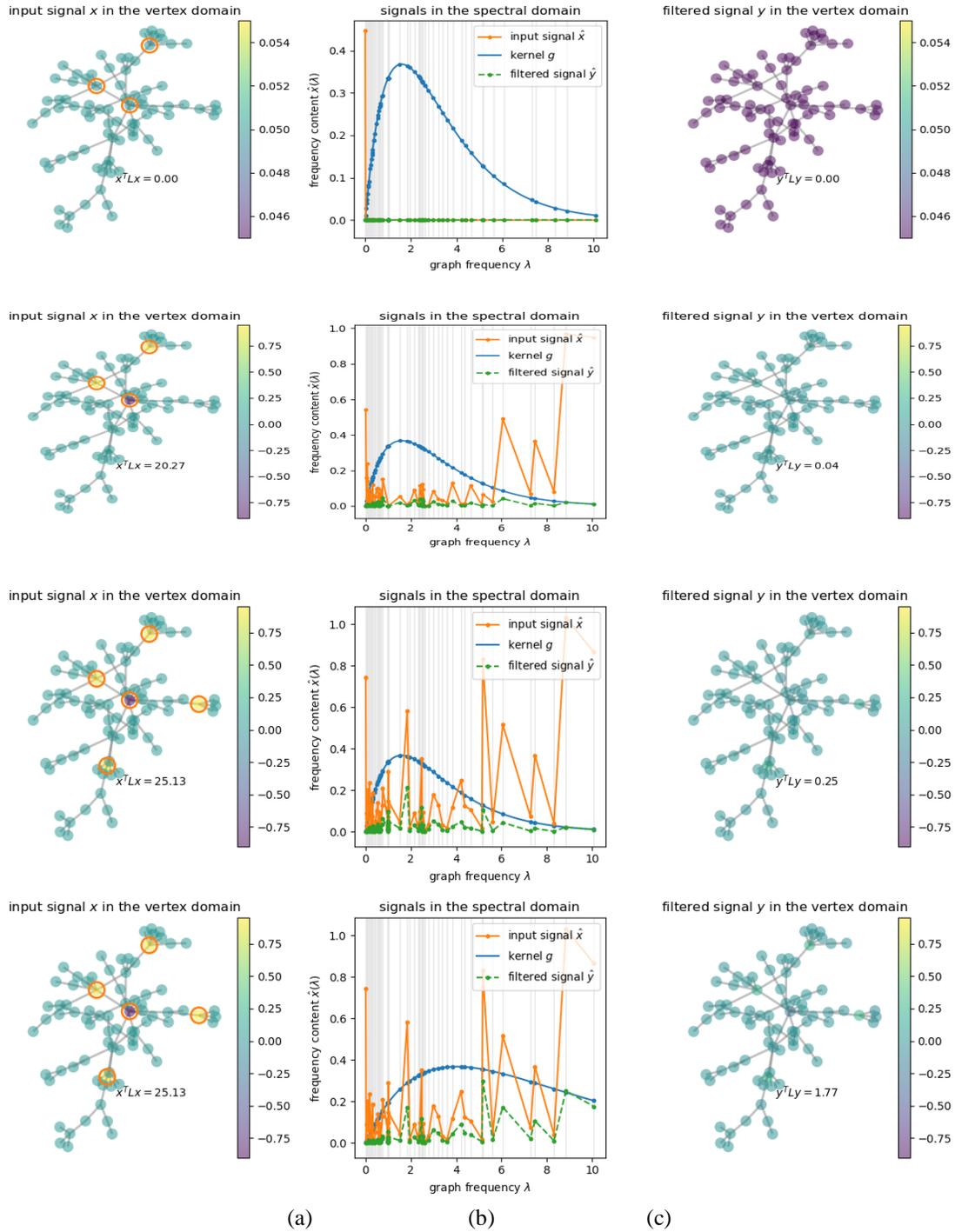

Figure 3. (a) Input signal in the vertex domain with anomaly nodes (b) The result of graph signals band-pass filtering (c) Filtered signal in the vertex domain.

Through observation of the above graph, it is noted that the band-pass filter applied to the graph signal effectively captures the frequency changes caused by anomalous nodes to a certain extent. However, the filtering outcome induces signal smoothness, consequently diminishing the discriminability of anomaly nodes to a certain degree.

### III. METHODOLOGY

A. Background: ChebConv [20]

ChebConv model is an inheritable outcome of the Spectral Convolutional Neural Network and a predefined diagonal convolutional kernel is defined as:

$$g_\theta = diag(\{\theta_i\}_{i=0}^{N-1}) \tag{5}$$

The result of convolving signal $x$ with $g_\theta$ is represented as:

$$y = U g_\theta U^T x = \begin{pmatrix} \theta_0 u_0 u_0^T + \theta_1 u_1 u_1^T \cdots \\ \theta_{N-1} u_{N-1} u_{N-1}^T \end{pmatrix} \tag{6}$$

To alleviate the high computational complexity associated with the need for eigen decomposition of the Laplacian matrix in the model proposed by Spectral GNN [21], Defferrard et al., [20] introduced a parameterized convolution kernel:

$$g_\theta(\Lambda) = \sum_{k=0}^{K-1} \theta_k \Lambda^k \quad \theta \in \mathbb{R}^K \tag{7}$$

First, the signal $x$ undergoes a graph Fourier transform. Filtering is then performed by multiplying the convolution kernel with the signal $x$ in the frequency domain. After filtering, an inverse transform is applied to convert the result back to the spatial domain, yielding the filtered signal $y$. The signal $y$ can be represented as follows:

$$\begin{aligned} y &= U g_\theta U^T x \\ &= U \sum_{k=0}^{K-1} \theta_k \Lambda^k U^T x \end{aligned} \tag{8}$$

The equation can be expanded as follows:

$$= \begin{Bmatrix} \theta_0 I + \theta_1 (\lambda_0 u_0 u_0^T + \cdots + \lambda_{n-1} u_{n-1} u_{n-1}^T) + \cdots \\ + \alpha_n^{k-1} (\lambda_0^{K-1} u_0 u_0^T + \cdots + \lambda_{n-1}^{K-1} u_{n-1} u_{n-1}^T) \end{Bmatrix}$$

>

$$= (\theta_0 I + \theta_1 L + \theta_1 L^2 + \cdots + \theta_{K-1} L^{K-1})x \tag{9}$$

Delving into the spectral properties of the ChebConv-based model, according to the expanded formula, it is evident that the convolution kernels employed in this approach effectively serve as high-pass filters. These filters amplify high-frequency signals while attenuating low-frequency components. which renders the model less suitable for graph node classification problems that rely on the principle of local smoothness.

In contrast, the graph anomaly node detection task, the focus of this study, presents a unique set of challenges. The sparsity of anomaly nodes, often surrounded by normal nodes, makes their identification particularly difficult. These anomaly nodes tend to be "submerged" within the normal nodes, defying the assumptions of local smoothness. Consequently, their spectral representation exhibits high-frequency characteristics. This alignment with the strengths of high-pass filters makes ChebConv an ideal choice for capturing and isolating anomaly nodes in this context.

The following shows the situation where anomalous nodes are embedded among normal nodes, and the result after applying the high-pass filter.

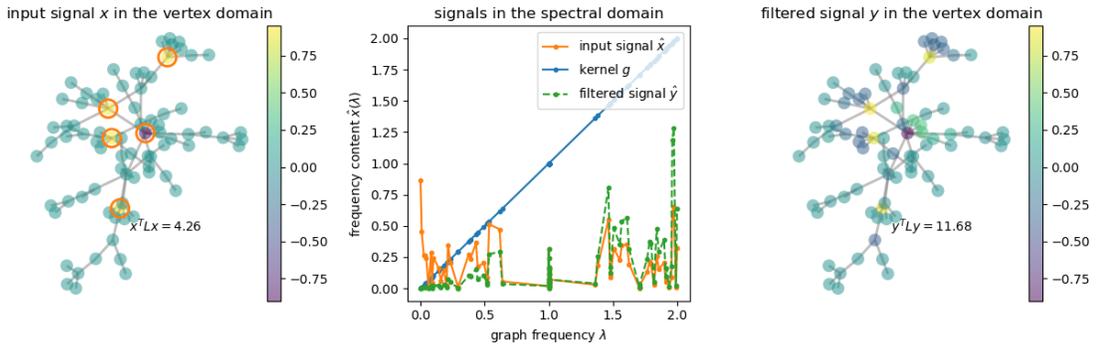

(a) $K$ is represented by 1

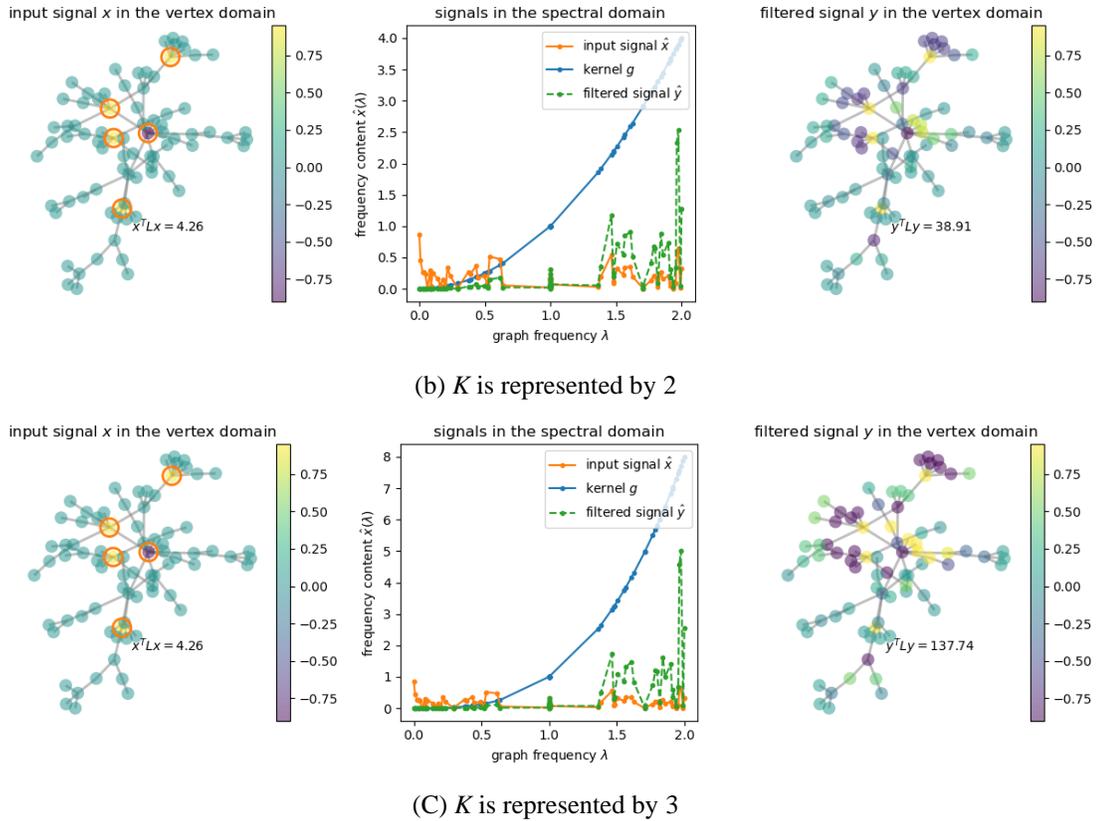

(b) *K* is represented by 2

(C) *K* is represented by 3

Figure 4．High-pass filter effect on the graph with varying values of parameter *K*.

High-pass filter can effectively preserve or enhance the high-frequency components，enhancing the distinctiveness of anomalous nodes. Based on the result shown in Figure 4, ChebConv model with high-pass filter functionality is utilized in this paper for graph anomaly node detection.

B. The isolated nodes in the graph are separated.

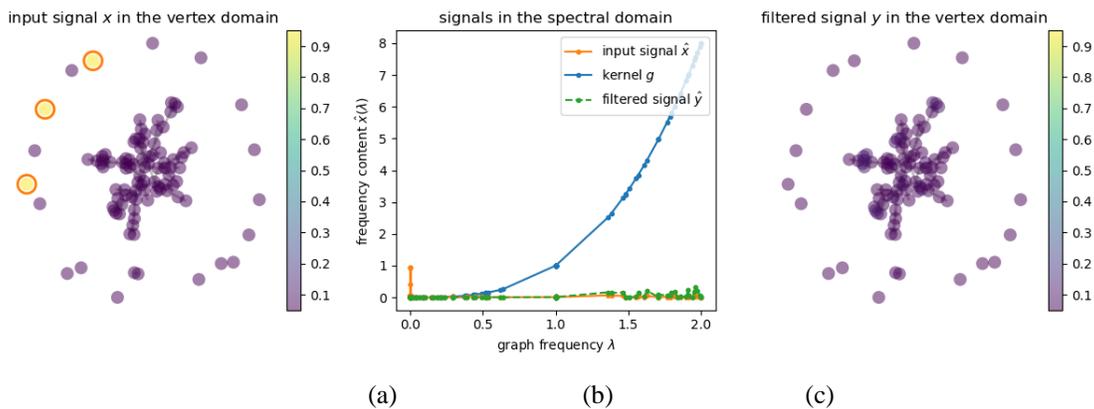

(a)　　　　　　　　　　(b)　　　　　　　　　　(c)

>

(1) anomalous nodes exist exclusively among the isolated nodes.

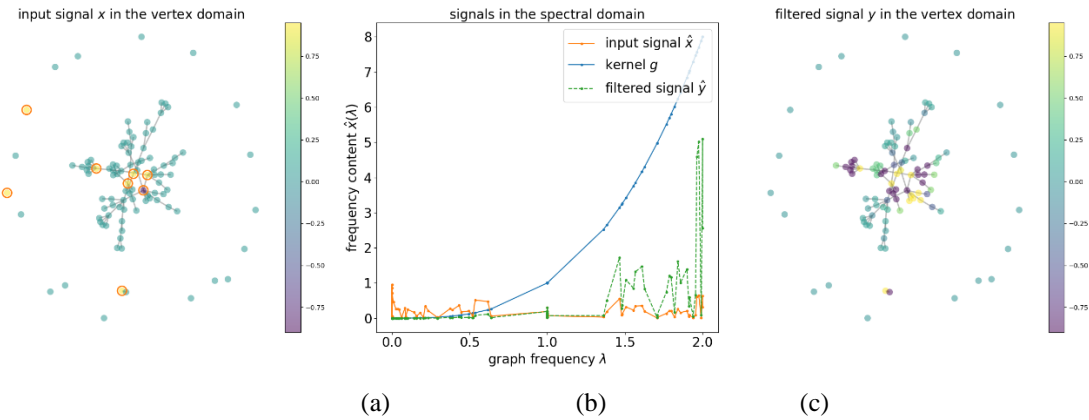

(a)          (b)          (c)

(2) Anomalous nodes are distributed throughout the graph.

Figure 5. Graph high pass filtering with isolated nodes and small connected components. (a) Input signal with anomaly nodes (b) The result of graph signals high-pass filtering (c) Filtered signal.

As shown in Figure 5 (1), anomalous nodes are only present among isolated nodes. The features of these nodes contain high-frequency components induced by the spectral graph and are lower than the Direct current (DC) component. After high-pass filtering, the node features become smoothed, reducing the distinguishability of the node features. In Figure 5 (2), anomalous nodes are distributed throughout the graph. After high-pass filtering, the features of isolated nodes also become smoothed, reducing the distinguishability of the node features.

For the datasets YelpChi and Amazon, the heterogeneous graphs contain a significant number of isolated nodes. As shown in Figure 2, the visualizations of Amazon-U-P-U (1720 nodes, total nodes 11,944) and YelpChi-R-U-R (22,123 nodes, total nodes 45,954) exhibit numerous isolated nodes and small connected components surrounding the central large, connected component. Among these isolated nodes, there are a certain number of anomalous nodes. which pose challenges to effective node representation learning

>

through graph neural networks. In this paper, these isolated nodes are separated from the graph and their features are learned using MLP layers.

C. A fusion model based on high pass filtered ChebConv and feature learning.

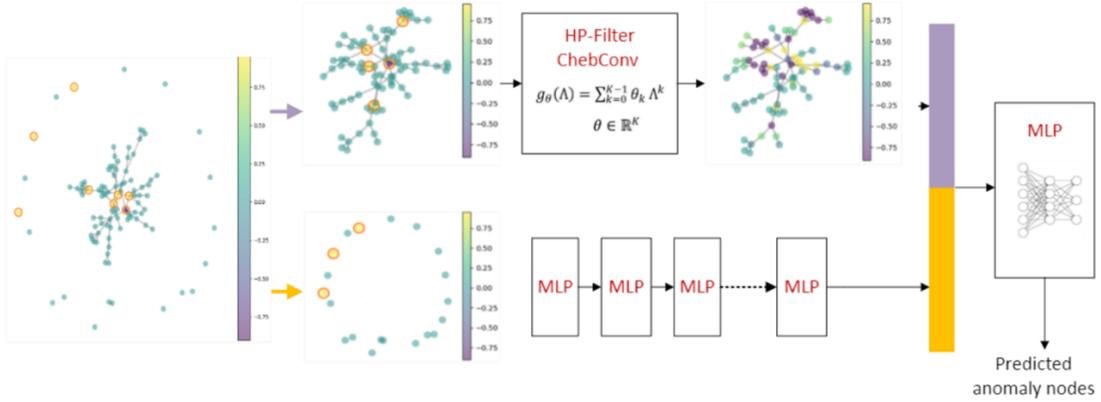

Figure 6. model architecture of HP-GCN_GAD

The model first segments the graph into isolated nodes and nodes within connected subgraphs. High-pass filtering is then applied to the connected subgraph using the ChebConv model. Following this, Multilayer Perceptions (MLPs) are used to generate the hidden representations of each node within the connected subgraph. Isolated nodes, on the other hand, learn their features solely through MLPs. Finally, the representations of both groups of nodes are merged, and another MLP is employed to predict the node categories.

## 5. RESULTS

In this section, we conduct empirical evaluations on real-world datasets and present the outcomes of our proposed models alongside several state-of-the-art baselines to demonstrate the efficacy of our approach. Specifically, our primary objective is to address the following research questions:

RQ1: How does the performance of the proposed model stack up against homophilous, heterophilous, and leading Graph Anomaly Detection (GAD) methods?

RQ2: Does using a graph convolutional neural network based on high-pass filters effectively enhance the detection of anomalous nodes in graphs? Additionally, what is the impact of the parameter $K$ of the high-pass filter on its performance?

RQ3: Can feature learning effectively detect anomalous nodes in isolated nodes and small connected components? Additionally, how does the number of layers in feature learning affect performance?

A. Experimental Setup

1. Datasets

**Table1. Statistics of Datasets**

| Dataset | #Nodes | #edges | Positive-Negative ratio | #features | Relation | # Edges |
|---|---|---|---|---|---|---|
| YelpChi | 45954 | 3,846,979 | 1:5.9 | 32 | R-U-R | 49,315 |
|  |  |  |  |  | R-S-R | 3,402,743 |
|  |  |  |  |  | R-T-R | 573,616 |
| Amazon | 11944 | 4,398,392 | 1:10.5 | 25 | U-P-U | 175,608 |
|  |  |  |  |  | U-S-U | 3,566,479 |
|  |  |  |  |  | U-V-U | 1,036,737 |
| T-Finance | 39357 | 21,222,543 | 1:21.8 | 10 | - | - |
| T-Social | 5,781,065 | 73,105,508 | 1:33.2 | 10 | - | - |

This paper conducts comparative experiments on four large datasets, with statistical information provided in Table 1. The YelpChi dataset consists of hotel and restaurant reviews, which are labeled by YelpChi as either positive (spam) or negative (legitimate). This dataset is utilized for detecting anomalous reviews. In the graph, reviews are represented as nodes, and three types of edges are defined:

R-U-R Edges: Connect reviews posted by the same user.

R-S-R Edges: Link reviews of the same product that share identical star ratings.

R-T-R Edges: Connect reviews of the same product posted within the same month.

In the Amazon Musical Instruments dataset, Users are represented as nodes in the graph. The graph incorporates different types of edges as follows:

U-P-U Edges: Connect users who have reviewed at least one common product.

U-S-U Edges: Link users who have given the same star rating to at least one product within a week.

U-V-U Edges: Connect users whose review texts exhibit top 5% similarity based on TF_IDF scores.

T-Finance dataset focuses on identifying anomalous accounts within transaction networks, where each user is represented as an account characterized by 10-dimensional features such as registration duration, login frequency, and interaction rates. Edges in this dataset denote transactional connections between pairs of accounts. Nodes are labeled as anomalies by human experts if they exhibit characteristics typical of activities such as online gambling, fraud, and money laundering.

On the other hand, the T-Social dataset aims to detect anomalous accounts within social networks. It shares node annotations and features with the T-Finance dataset, implying that similar criteria and methods are used for anomaly detection. In the T-Social dataset, edges are established between two nodes if their friendship has persisted for more than three months. Notably, the scale of T-Social is immense, outsizing datasets like YelpChi and Amazon by a factor of 100.

**2. baselines**

Our algorithm is evaluated against three sets of baseline models. The first set consists of general Graph Neural Network (GNN) models, encompassing GCN [22], GIN [23], GAT [24]. The second set focuses on neighborhood reconstruction and the detection of various types of anomalies, such as NWR-GAE [25], GAD_NR [26]. The third set includes state-of-the-art methods for graph-based anomaly detection, CAREGNN [11], PC-GNN [16] , GPRGNN [27], BWGNN [18], GHRN [9].

In the context of dealing with multi-relational graphs such as those found in YelpChi and Amazon, the approach to handle heterogeneity involves two distinct methods, inspired by BWGNN:

1. HP_GCN_GAD (homo): In this method, all types of edges in the graph are treated as identical. This means that the model does not distinguish between different types of relationships when learning the graph's structure and features. This simplification can make the model less complex and faster to train, but it might lose some information by ignoring the heterogeneity of the graph.

2. HP_GCN_GAD (hetero): This method preserves the distinction between different types of relationships. Graph filtering is conducted separately for each type of relation, and then the results are combined using a maximum pooling operation. This approach aims to capture the unique characteristics of each relationship type, potentially leading to a more nuanced and accurate model, albeit at the cost of increased complexity and computational demands.

The training methodology for the HP_GCN_GAD models involve several detailed settings and procedures. Here is a structured summary of the approach:

1. Training Setup:

- Epochs: All models were trained for 100 epochs.

- Optimizer: Adam optimizer was used with a learning rate of 0.01.

- Model Selection: The model with the best Macro-F1 score on the validation set

was saved.

2. Datasets and Parameters:

- Datasets: YelpChi, Amazon, T-Finance, and T-Social.

- Representation and Hidden Layer Dimensions (h): Set to 64 for all models.

- ChebConv Power (K): Set to 3 for all datasets, except T-Social where it was set to 6.

3. Aggregation Function:

- AGG(·): Used concatenation in HP_GCN_GAD.

4. Training Scenarios:

- Supervised Scenario:

  o Training Proportion: 40% of the data.

- Semi-Supervised Scenario:

  o Training Proportion: 1% of the data.

  o Data Split for Validation and Testing: The remaining data was split in a 1:2 ratio for validation and testing.

This setup ensures a consistent training procedure across different datasets and scenarios, facilitating a robust evaluation of the model's performance under varying conditions.

### 3. Metrics

To thoroughly assess the performance of the compared methods, we use two well-established metrics: F1-macro and AUC. The F1-macro score, which is the unweighted average of the F1-scores for both classes (normal and anomaly), is particularly effective for imbalanced datasets where the number of normal data points may greatly exceed anomalies. The AUC (Area Under the Receiver Operating Characteristic (ROC) Curve), as defined by Davis and Goadrich (2006) [28], offers a complementary perspective by evaluating the model's ability to distinguish between normal and anomalous data points across all possible classification thresholds.

**4. Implementation Details**

(1). Performance Comparison

The performance of different models across various datasets is summarized in Table 2 and Table 3. Below are the key observations and findings:

➢ Overall Performance:

- HP-GNN_GAD typically attains the superior performance across most datasets.

- On Amazon (1%): PC-GNN emerges as the top performer with the highest AUC score, and BHomo-GNRN excels in achieving the best F1-Macro score.

➢ Multi-Relation Graphs:

- YelpChi: HP-GNN_GAD (Hetero) outperforms other models.

- Amazon: HP-GNN_GAD (Homo) shows better performance compared to its hetero counterpart.

➢ Comparison with State-of-the-Art Methods:

- CAREGNN, PC-GNN, BWGNN, and GNRN: these represent four state-of-the-art methods for anomaly detection leveraging graph-based approaches.

- HP-GNN_GAD vs. BWGNN (hetero) on YelpChi (40%): HP-GNN_GAD shows

196an absolute improvement of 8.18% in F1-Macro and 5.68% in AUC.

- HP-GNN_GAD vs. GAD_NR on T-Social (40%): HP-GNN_GAD demonstrates a significant improvement with an 18.77% increase in F1-Macro and a 4.65% increase in AUC, highlighting both its superiority and scalability on large graphs.

➢ Performance evaluation of General GNN Models:

- When comparing GCN, GAT and GIN, GCN often exhibits the least satisfactory performance. This aligns with the observation that a low-pass filter is inadequate for accurately distinguishing anomalies.

➢ Neighborhood Reconstruction and Anomaly Detection:

- NWR-GAE and GAD_NR: These models focus on neighborhood reconstruction and detecting various types of anomalies.

- T-Social (40%): Here, performance improvements are noted with NWR-GAE and GAD_NR, showing a 5.1% increase in F1-Macro and a 2.15% increase in AUC compared to BWGNN.

In summary, HP-GNN_GAD consistently outperforms several state-of-the-art methods, particularly excelling on datasets with large graphs and multiple relations, showcasing its effectiveness and Scalability in graph-based anomaly detection.

Table 2. Experimental results with 40% training ratios.

| Dataset Metric | YelpChi(40%) F1-Macro  AUC | Amazon(40%) F1-Macro  AUC | T-finance(40%) F1-Macro  AUC | T-Social(40%) F1-Macro  AUC |
|---|---|---|---|---|
| GCN | 53.24  54.28 | 67.75  83.08 | 82.04  86.58 | 55.51  65.36 |
| GAT | 52.33  53.97 | 83.48  89.80 | 77.37  79.51 | 63.73  77.54 |
| GIN | 64.12  75.78 | 69.36  78.38 | 60.21  73.46 | 57.65  83.46 |
| CAREGNN(homo) | 66.55  79.50 | 92.70  95.99 | 86.24  91.89 | -  - |
| CAREGNN(hetero) | 69.00  82.50 | 89.99  92.59 | -  - | -  - |
| PC-GNN(homo) | 61.29  78.12 | 87.29  95.20 | 46.47  93.37 | 49.73  90.18 |
| PC-GNN(hetero) | 67.75  86.29 | 88.15  96.48 | -  - | -  - |

>

| | | | | | | | | |
|---|---|---|---|---|---|---|---|---|
| GPRGNN | 65.14 | 77.73 | 89.14 | 92.97 | 78.46 | 89.60 | 65.79 | 86.54 |
| BWGNN(homo) | 68.41 | 81.70 | 91.50 | 97.08 | 89.93 | 94.66 | 71.06 | 92.14 |
| BWGNN(hetero) | 77.44 | 90.42 | 91.81 | 97.44 | - | - | - | - |
| BHomo-GHRN | 70.38 | 83.59 | 92.27 | 97.61 | 89.09 | 96.03 | 69.29 | 90.82 |
| BHetero-GHRN | 76.84 | 90.42 | 91.65 | 97.34 | - | - | - | - |
| NWR-GAE | 56.46 | 61.22 | 57.70 | 72.06 | 51.18 | 55.96 | 66.37 | 78.26 |
| GAD_NR | 54.30 | 57.84 | 71.96 | 84.84 | 52.67 | 75.36 | 76.16 | 94.29 |
| Ours(homo) | 70.29 | 83.71 | **92.25** | **98.16** | **91.13** | **96.46** | **94.93** | **98.94** |
| Ours(hetero) | **85.62** | **96.10** | 91.58 | 97.19 | - | - | - | - |

Table 3. Experimental results with 1% training ratios.

| Dataset Metric | YelpChi(1%) F1-Macro AUC | | Amazon(1%) F1-Macro AUC | | T-finance(1%) F1-Macro AUC | | T-Social(1%) F1-Macro AUC | |
|---|---|---|---|---|---|---|---|---|
| GCN | 53.60 | 57.90 | 66.70 | 81.73 | 81.64 | 87.36 | 53.39 | 63.49 |
| GAT | 52.79 | 54.43 | 59.08 | 66.80 | 76.78 | 76.40 | 59.90 | 73.04 |
| GIN | 62.56 | 73.86 | 65.24 | 75.81 | 67.57 | 83.06 | 56.09 | 72.67 |
| CAREGNN(homo) | 62.20 | 72.53 | 89.86 | 90.25 | 85.38 | 92.92 | - | - |
| CAREGNN(hetero) | 63.75 | 75.32 | 85.27 | 90.14 | - | - | - | - |
| PC-GNN(homo) | 60.91 | 74.10 | 51.64 | **91.52** | 56.25 | 93.16 | 36.56 | 88.54 |
| PC-GNN (hetero) | 60.37 | 81.46 | 74.76 | 88.56 | - | - | - | - |
| GPRGNN | 61.82 | 71.66 | 82.07 | 87.52 | 70.61 | 88.01 | 81.29 | 95.08 |
| BWGNN(homo) | 61.55 | 72.37 | 89.64 | 89.49 | 84.63 | 89.42 | 68.67 | 87.28 |
| BWGNN(hetero) | 65.62 | 71.72 | 85.37 | 86.95 | - | - | - | - |
| BHomo-GHRN | 61.85 | 72.00 | **91.55** | 87.99 | 86.91 | 91.92 | 64.56 | 87.35 |
| BHetero-GHRN | 65.00 | 74.86 | 81.75 | 87.65 | - | - | - | - |
| NWR-GAE | 56.07 | 60.21 | 63.79 | 75.76 | 52.11 | 57.77 | 66.42 | 78.22 |
| GAD_NR | 52.96 | 54.92 | 66.65 | 82.13 | 48.75 | 57.02 | 76.55 | 94.91 |
| Ours(homo) | 61.78 | 72.36 | 89.94 | 91.24 | **88.21** | **93.23** | **94.39** | **98.58** |
| Ours(hetero) | **78.46** | **86.16** | 86.58 | 87.15 | - | - | - | - |

B. Sensitivity Analysis

1. The Power K in ChebConv model

The parameter K plays a critical role as a hyperparameter in Chebyshev. Different values of K correspond to different graph high-pass filters.

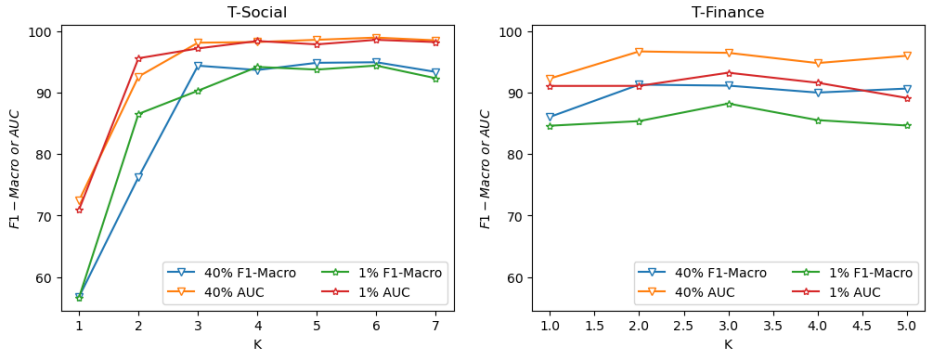

Figure 7. Our method performance with varying values of parameter K.

As shown in Figure 7 (right), on the T-Finance dataset, when K is set to 2 and the training rate is 40%, both the F1-Macro and AUC metrics reach their maximum values, which are 91.29% and 96.69%, respectively. When K is set to 3 and the training rate is 1%, the maximum values of these two metrics are 88.21% and 93.23%, respectively. For the T-Social dataset, As shown in Figure 7 (left), both the F1-Macro and AUC metrics exhibit a positive correlation with the value of K. When the training rate is 40%, the F1-Macro shows a slight fluctuation at K = 4. However, beyond K = 3, both metrics tend to stabilize and exhibit a slight improvement, reaching maximum values of 94.93% and 98.94% respectively, at K=6. When the training rate is 1%, both the F1-Macro and AUC metrics achieve their maximum values at K = 6, which are 94.39% and 98.58%, respectively. Furthermore, as the value of K increases, time consumption also increases, as illustrated in Figure 8.

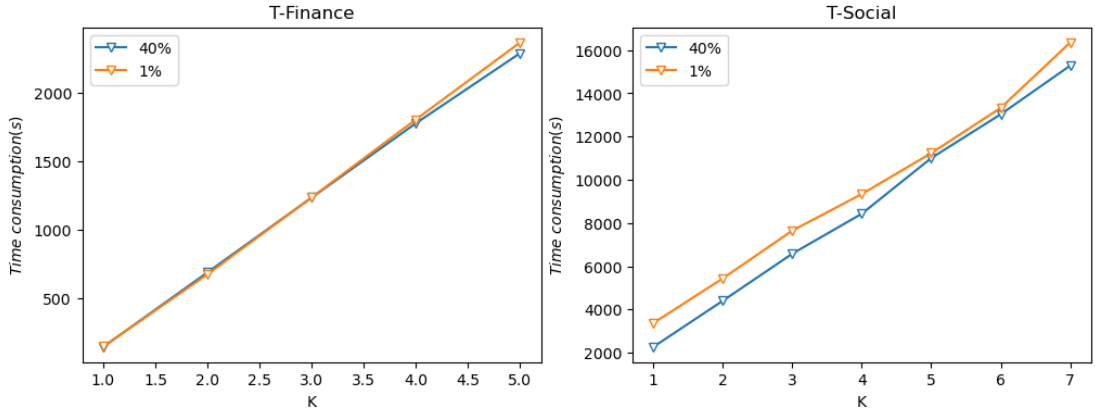

Figure 8. Our method time consumption with varying values of the parameter K.

## 5. CONCLUSION

This paper introduces a novel method to address the graph anomaly detection problem, where the distribution of anomalous and normal nodes is unbalanced, and anomalous nodes are typically embedded within normal nodes. The method utilizes high-pass filtering within the spectral graph convolution network to detect anomalous nodes. Additionally, to tackle the challenge posed by independent nodes in the graph, which cannot be effectively learned by graph neural network models, the independent nodes are separated from the original graph and learned using a Multi-Layer Perceptron (MLP). The paper presents experimental comparisons with 10 advanced models for detecting graph anomalies across four datasets. Experimental results demonstrate the effectiveness of employing high-pass filters for graph anomaly detection (HP-GCN) on all four datasets.

The algorithm proposed in this paper is applicable to two types of graph anomaly detection scenarios:

1. Graphs with a large number of independent nodes: In the case of the YelpChi dataset, although the anomaly-to-normal node ratio is relatively high (1:5.9), the large number of independent nodes in the graph (45,954 nodes, of which 22,123 are isolated) presents a unique challenge. Independent nodes cannot be effectively learned by graph convolutional neural network, and graph filters struggle to filter out the associated noise. By separating these independent nodes and applying MLP, more effective learning of node representations is achieved, resulting in an accuracy of 98.16%.

2. Low anomaly node proportion graph anomaly detection: In scenarios where the ratio of anomaly nodes to normal nodes is low, the algorithm shows slight performance improvements on the Amazon (anomaly-to-normal node ratio: 1:10.5) and T-Finance (anomaly-to-normal node ratio: 1:21.8) datasets. However, for the T-Social dataset, which has an even lower anomaly-to-normal node ratio (1:33.2), anomaly detection accuracy is significantly enhanced to 98.94%, compared to 94.29% accuracy by the existing best algorithm [26]. Additionally, T-Social is a large-scale dataset with over 5 million nodes, and the considerable performance improvement on this dataset further demonstrates the scalability of the proposed algorithm.

This work expands the understanding of graph anomaly detection in the spectral domain. It offers a new approach for anomaly detection in graphs using Graph Neural Network (GNN). Future research should aim to refine and extend these techniques to address a broader range of applications and challenges in the field of graph anomaly detection. Future work should aim to refine and extend these techniques to tackle a broader range of applications and challenges in the field of graph anomaly detection.

## ACKNOWLEDGMENT

This study was supported by the Hainan Provincial Natural Science Foundation of China (No.621MS058), the school-level Science Foundation of Sanya university (No. USYZD22-07), the specific research fund of The Innovation Platform for Academician of Hainan Province (No. YSPTZX202144, No. YSPTZX202145).

## AUTHOR STATEMENT

**Li Shelei:** Software, Methodology, Writing-Original draft preparation. **Tan Yong Chai:** Conceptualization, Software.

## DECLARATION OF COMPETING INTEREST

The authors declare that they have no conflicts of interest to report regarding the present study.

\>

## REFERENCES

[1] X. Ma, J. Wu, S. Xue, et al., "A Comprehensive Survey on Graph Anomaly Detection With Deep Learning," IEEE Trans. Knowl. Data Eng., vol. 35, no. 12, pp. 12012–12038, Dec. 2022, doi: 10.1109/TKDE.2021.3118815.

[2] M. Zakariah and A. S. Almazyad, "Anomaly Detection for IOT Systems Using Active Learning," Appl. Sci., vol. 13, no. 21, p. 12029, Nov. 2023, doi: 10.3390/app132112029.

[3] L. Wu, X. Hu, F. Morstatter, et al., "Adaptive Spammer Detection with Sparse Group Modeling," Proc. Int. AAAI Conf. Web Soc. Media, vol. 11, no. 1, pp. 319–326, May 2017, doi: 10.1609/icwsm.v11i1.14887.

[4] Y. Wu, H.-N. Dai, and H. Tang, "Graph Neural Networks for Anomaly Detection in Industrial Internet of Things," IEEE Internet Things J., vol. 9, no. 12, pp. 9214–9231, Jun. 2022, doi: 10.1109/JIOT.2021.3094295.

[5] L. Xiao, X. Yang, and X. Yang, "A graph neural network-based bearing fault detection method," Sci. Rep., vol. 13, no. 1, p. 5286, Mar. 2023, doi: 10.1038/s41598-023-32369-y.

[6] X. Huang, Y. Yang, Y. Wang, et al., "DGraph: A Large-Scale Financial Dataset for Graph Anomaly Detection," in Advances in Neural Information Processing Systems 35 (NeurIPS 2022), 2022. [Online]. Available: https://www.ic3.gov/Media/PDF/AnnualReport/2020_IC3Report.pdf

[7] N. Innan, A. Sawaika, A. Dhor, et al., "Financial fraud detection using quantum graph neural networks," Quantum Mach. Intell., vol. 6, no. 1, p. 7, Jun. 2024, doi: 10.1007/s42484-024-00143-6.

[8] G. Liu, J. Tang, Y. Tian, et al., "Graph Neural Network for Credit Card Fraud Detection," in 2021 International Conference on Cyber-Physical Social Intelligence (ICCSI), IEEE, Dec. 2021, pp. 1–6. doi: 10.1109/ICCSI53130.2021.9736204.

[9] Y. Gao, X. Wang, X. He, et al., "Addressing Heterophily in Graph Anomaly Detection: A Perspective of Graph Spectrum," in ACM Web Conference 2023 - Proceedings of the World Wide Web Conference, WWW 2023, Association for Computing Machinery, Inc, Apr. 2023, pp. 1528–1538. doi: 10.1145/3543507.3583268.

[10] D. Wang, Y. Qi, J. Lin, et al., "A Semi-Supervised Graph Attentive Network for Financial Fraud Detection," in 2019 IEEE International Conference on Data Mining (ICDM), IEEE, Nov. 2019, pp. 598–607. doi: 10.1109/ICDM.2019.00070.

[11] Y. Dou, Z. Liu, L. Sun, et al., "Enhancing Graph Neural Network-based Fraud Detectors against Camouflaged Fraudsters," in Proceedings of the 29th ACM International Conference on Information & Knowledge Management, New York, NY, USA: ACM, Oct. 2020, pp. 315–324. doi: 10.1145/3340531.3411903.

[12] A. Deng and B. Hooi, "Graph Neural Network-Based Anomaly Detection in Multivariate Time Series," Proc. AAAI Conf. Artif. Intell., vol. 35, no. 5, pp. 4027–4035, May 2021, doi: 10.1609/aaai.v35i5.16523.

[13] Y. Liu, Z. Li, S. Pan, et al., "Anomaly Detection on Attributed Networks via Contrastive Self-Supervised Learning," IEEE Trans. Neural Networks Learn. Syst., vol. 33, no. 6, pp. 2378–2392, Jun. 2022, doi: 10.1109/TNNLS.2021.3068344.

[14] F. Shi and C. Zhao, "Enhancing financial fraud detection with hierarchical graph attention networks: A study on integrating local and extensive structural information," Financ. Res. Lett., vol. 58, Dec. 2023, doi: 10.1016/j.frl.2023.104458.

[15] Z. Liu, Y. Dou, P. S. Yu, et al., "Alleviating the Inconsistency Problem of Applying Graph Neural Network to Fraud Detection," SIGIR 2020 - Proc. 43rd Int. ACM SIGIR Conf. Res. Dev. Inf. Retr., pp. 1569–1572, May 2020, doi: 10.1145/3397271.3401253.

[16] Y. Liu, X. Ao, Z. Qin, et al., "Pick and choose: A GNN-based imbalanced learning approach for fraud detection," in The Web Conference 2021 - Proceedings of the World Wide Web Conference, WWW 2021, Association for Computing Machinery, Inc, Apr. 2021, pp. 3168–3177. doi: 10.1145/3442381.3449989.

[17] Y. Gao, X. Wang, X. He, et al., "Alleviating Structural Distribution Shift in Graph Anomaly Detection," in Proceedings of the Sixteenth ACM International Conference on Web Search and Data Mining, New York, NY, USA: ACM, Feb. 2023, pp. 357–365. doi: 10.1145/3539597.3570377.

[18] J. Tang, J. Li, Z. Gao, et al., "Rethinking Graph Neural Networks for Anomaly Detection," in Proceedings of the 39th International Conference on Machine Learning, ICML, 2022, pp. 21076–21089. doi: https://arxiv.org/abs/2205.15508.

[19] B. Xu, H. Shen, Q. Cao, et al., "Graph Wavelet Neural Network," in ICLR, ICLR, 2019.

[20] M. Defferrard, X. Bresson, and P. Vandergheynst, "Convolutional Neural Networks on Graphs with Fast Localized Spectral Filtering," Thirtieth Annu. Conf. Neural Inf. Process. Syst., Jun. 2016, [Online]. Available: http://arxiv.org/abs/1606.09375.

[21] J. Bruna, W. Zaremba, A. Szlam, et al., "Spectral Networks and Locally Connected Networks on Graphs," 2nd Int. Conf. Learn. Represent., Dec. 2014, [Online]. Available: http://arxiv.org/abs/1312.6203.

[22] T. N. Kipf and M. Welling, "Semi-Supervised Classification with Graph Convolutional Networks," Sep. 2016. [Online]. Available: http://arxiv.org/abs/1609.02907.

[23] X. Wang and M. Zhang, "How Powerful are Spectral Graph Neural Networks," in Proceedings of the 39th International Conference on Machine Learning, May 2022, pp. 23341–23362. [Online]. Available: http://arxiv.org/abs/2205.11172

[24] Veličković Petar, "Graph Attention networks," 2018. doi: 10.48550/arXiv.1710.10903.

[25] M. Tang, C. Yang, and P. Li, "Graph Auto-Encoder Via Neighborhood Wasserstein Reconstruction," 2022. [Online]. Available: https://github.com/mtang724/NWR-GAE.

[26] A. Roy, J. Shu, J. Li, et al., "GAD-NR: Graph Anomaly Detection via Neighborhood Reconstruction," in WSDM 2024 - Proceedings of the 17th ACM International Conference on Web Search and Data Mining, Association for Computing Machinery, Inc, Mar. 2024, pp. 576–585. doi: 10.1145/3616855.3635767.

[27] E. Chien, J. Peng, P. Li, et al., "Adaptive Universal Generalized PageRank Graph Neural Network," in ICLR, Jun. 2021. [Online]. Available: http://arxiv.org/abs/2006.07988

[28] J. Davis and M. Goadrich, "The relationship between Precision-Recall and ROC curves," in Proceedings of the 23rd international conference on Machine learning - ICML '06, 2006, pp. 233–240. doi: 10.1145/1143844.1143874.

**Li Shelei** is a professor with School of Information & Intelligence Engineering. Sanya University. Currently, she is a PhD candidate at SEGi university.

**Tan Yong Chai** is an associate professor with the Faculty of Engineering, Building Environment and information Technolog**y,** SEGi university. He is PHD supervisor.